\documentclass{article}

\usepackage{times}
\usepackage{graphicx} 
\usepackage{subfigure} 

\usepackage{natbib}

\usepackage{algorithm}
\usepackage{algorithmic}

\usepackage{hyperref}

\usepackage[accepted]{icml2017}

\icmltitlerunning{Meta learning Framework for Automated Driving }

\begin{document} 

\twocolumn[
\icmltitle{Meta learning Framework for Automated Driving }



\icmlsetsymbol{equal}{*}

\begin{icmlauthorlist}
\icmlauthor{Ahmad El Sallab}{Sallab}
\icmlauthor{Mahmoud Saeed}{equal,Mahmoud}
\icmlauthor{Omar Abdel Tawab}{equal,Omar}
\icmlauthor{Mohammed Abdou}{Abdou}
\end{icmlauthorlist}

\icmlaffiliation{Sallab}{Ahmad El Sallab is a Senior Expert at Valeo, ahmad.el-sallab@valeo.com}
\icmlaffiliation{Mahmoud}{Mahmoud Saeed is an Intern at Valeo, mahmoud.saeed.ext@valeo.com}
\icmlaffiliation{Omar}{Omar Abdel Tawab is an Intern at Valeo, omar.abdeltawab.ext@valeo.com}
\icmlaffiliation{Abdou}{Mohammed Abdou is a Researcher at Valeo, mohammed.abdou@valeo.com}

\icmlcorrespondingauthor{Ahmad El Sallab}{ahmad.el-sallab@valeo.com}

\icmlkeywords{boring formatting information, machine learning, ICML}

\vskip 0.3in
]



\printAffiliationsAndNotice{\icmlEqualContribution} 

\begin{abstract} 
The success of automated driving deployment is highly depending on the ability to develop an efficient and safe driving policy. The problem is well formulated under the framework of optimal control as a cost optimization problem. Model based solutions using traditional planning are efficient, but require the knowledge of the environment model. On the other hand, model free solutions suffer sample inefficiency and require too many interactions with the environment, which is infeasible in practice. Methods under the Reinforcement Learning framework usually require the notion of a reward function, which is not available in the real world. Imitation learning helps in improving sample efficiency by introducing prior knowledge obtained from the demonstrated behavior, on the risk of exact behavior cloning without generalizing to unseen environments. In this paper we propose a Meta learning framework, based on data set aggregation, to improve generalization of imitation learning algorithms. Under the proposed framework, we propose MetaDAgger, a novel algorithm which tackles the generalization issues in traditional imitation learning.  We use The Open Race Car Simulator (TORCS) to test our algorithm. Results on unseen test tracks show significant improvement over traditional imitation learning algorithms, improving the learning time and sample efficiency in the same time. The results are also supported by visualization of the learnt features to prove generalization of the captured details.
\end{abstract} 

\section{Introduction}
\label{submission}

Automated driving development has radically changed during the past few years, driven by advances in Artificial Intelligence (AI), and specifically Deep Learning (DL). 
Developing an efficient and safe driving policy is in the heart of reaching high level of autonomy in a robot car. Traditional methods are driven with rule based approaches \cite{le2006review}\cite{pasquier2001fuzzylot}, while significant advancement in the field is driven by learning approaches \cite{sallab2016end}\cite{el2017deep} \cite{bojarski2016end}. Reinforcement learning is the arm of AI which is concerned with solving the control problem based on learning while interacting with the environment.

Our motivation is to take the work in \cite{el2017deep} and \cite{sallab2016end} a step further towards real car deployment. The constraints of such task are: 1) No damage caused by interaction with real environment (which was possible in the game engines world), 2) Sample efficiency, so learning time has to be reasonable 3) Generalization, where the learnt policy of the driving agent should be able to capture the basics of the intuitive driving when deployed in unseen environments.

There are several approaches to find an efficient driving policy. Optimal control methods based on traditional planning require knowledge of an environment model. While model based approaches are gaining more popularity \cite{polydoros2017survey}, it is still hard to develop an efficient environment model especially for complex ones like urban and city scenarios. On the other hand, model free approaches \cite{Watkins1989a} \cite{sallab2016end} require the definition of a reward function, which only exist in simulated environments and game engines but not in real world.
Imitation learning is another successful approach, which might suffer the risk of de-generalization to unseen environments other than trained on. The most popular algorithms are based on data aggregation, which could be done through some hacks as in \cite{bojarski2016end}, or formally as in \cite{ross2011reduction}.

In this paper we propose a Meta learning framework for automated driving, to enforce transfer learning from one environment to another. Our hypothesis is that Meta learning will help capturing generic features without memorizing one specific environment. One proof of generalization is the visualization of which features the agent has learnt. If the learnt features are quite specific to the training environment, then it means that such agent will fail if deployed in another environment with different features. Under the Meta framework, we propose a novel algorithm; MetaDAgger, which is based on a Meta learning framework to aggregate data across different environments. To ensure generalization, we split the learning agent into two: 1) Meta learner, whose objective is to capture generic features not specific to an environment and 2) Low level learner, whose objective is to drive in a specific environment. The role of Meta learning is to smooth the learning performance across different environments. 

The two learners are trained using Convolutional Neural Networks (ConvNets) on two different data sets. The meta learner is trained on a data set of environments; where  a set of environments are kept for training, while another independent set of environments are kept for testing and are not allowed to alter the learning process. One the other hand, the Low learner is trained on a data set aggregated from each specific environment, in the form of state-action pairs. 
The link between both learners is done through continual lifelong learning, where the Low learner communicates its learning to the Meta learner whenever a switch from an environment to the other is undergone. In this way the learning is preserved across different environments, which enable the Meta learner to capture general features as proved by the features visualizations. It is worth noting that in \cite{sallab2016end} an interesting conclusion is that continual learning highly improves the learning curve, which is in line with the proposed metal learning framework proposed in this work.

The experimental setup is based on The Open Race Car Simulator (TORCS) \cite{wymann2000torcs}. The Meta dataset is formed of 19 tracks, with 10 training tracks and 9 testing tracks. Low learners are based on Convolutional Neural Networks (ConvNets) for policy learning. The results show significant advantage of MetaDAgger over DAgger on both training and testing tracks. MetaDAgger is not only improving generalization, but also significantly improving the sample efficiency and learning time over DAgger. The objective in each low level task is to keep the central lane. The demonstration in all tracks comes from a tradition PID controller, with access to the position of the ego car with respect to the left and right lanes. The control actions is applied to the steering wheel, with continuous actions output. The input states in all experiments are taken as the raw pixels information as provided by TORCS.

The rest of the paper is organized as follows; first we discuss the related work, then the Meta learning framework is described followed by the MetaDAgger algorithm. Then the experimental setup is described followed by the discussion of results and visualizations, and finally we conclude.

\section{Related Work}
\label{submission}

The problem of developing a driving policy is essentially formulated under the framework of optimal control. The solution of the optimization problem is easily found under the traditional planning framework only if an environment model exists, a requirement which is difficult to achieve in the real world of high way, urban or city driving. Model based approach is a wide topic that has been recently tackled in \cite{polydoros2017survey}.

On the other hand, model free approaches have undergone a huge advancement in the area of human level control, the Deep Q Net is a famous example of which \cite{MnihKavukcuogluSilverEtAl2015a} \cite{MnihKavukcuogluSilverEtAl2013a}. The success of model free Reinforcement Learning \cite{Sutton1988a} \cite{Watkins1989a} has reached the area Automated Driving \cite{Karavolos2013a} \cite{el2017deep} \cite{sallab2016end}. However, model free approaches require interacting with an environment through actions and rewards scheme. Such interaction usually requires a controlled environment such as simulators and game engines. The reward function is relatively easy acquired in such configuration of game engines. However, it is not clear how such reward function can be obtained in real world without causing too much damage. Moreover, model free approaches are known of their sample inefficiency, which is compensated by huge number of interactions with the simulated environment, which is again infeasible in real world. 

Another learning approach based on human demonstrations has been successfully deployed in automated driving \cite{bojarski2016end}.  Apprenticeship learning has been tackled in \cite{AbbeelNg2004} \cite{NgRussellothers2000a}, under the framework of Inverse Reinforcement Learning (IRL). IRL goal is to deduce the reward function the coach has been trying to achieve or maximize throughout the course of apprenticeship. Such hard goal is achieved on the expense of high complexity algorithm, involving two nested RL loops.

In order to tackle the problem of sample efficiency without knowing the environment models, supervised learning models are employed, where learning is based on demonstrated behavior to imitate (imitation learning). Learning from demonstrations has been approached as a supervised learning problem for automated driving in \cite{bojarski2016end}. The risk of such an approach is the lack of generalization, where supervised learning schemes could converge to exact behavior cloning. The challenge of getting such approaches to work lies in the ability to enrich the training data by introducing new unseen states from the demonstration. One solution is data aggregation, where the agent is able to extend his actions to further states than the demonstrated ones.  Some hacks have been proposed in \cite{bojarski2016end} to achieve data aggregation, where two cameras has been added in addition to the central camera to augment the training states with different situations supported with the correct action to take in each. A more formal approach has been formulated in the Dataset aggregation (DAgger) algorithm \cite{ross2011reduction}. DAgger algorithm is also extensible to include safety constraints, as in SafeDAgger \cite{zhang2016query}.

Transfer of experience is an essential component of the human learning process. Transfer of knowledge could be in the form of external coaching, demonstrations or apprenticeship, or it could take the form of self accumulated experience through lifelong, continual learning \cite{urgen1996simple}. Meta learning \cite{thrun2012learning} \cite{urgen1996simple} is another name of continual and transfer learning. Meta learning has also been used to search the best hyper parameters settings \cite{hochreiter2001learning} \cite{andrychowicz2016learning}. Continual learning fosters the transfer of experience across the life time of the agent. In addition, Meta learning provides a framework to transfer the knowledge acquired in one task to other tasks as well thanks to the captured information from task to task in the high level or Meta learner \cite{thrun2012learning}.

\section{Meta Learning Framework for Automated Driving}
\label{submission}

In this section the Meta Learning framework is presented $Figure$ \ref{Fig1}. The goal of the framework is to reach smooth and generalized performance over a set of environments\{${E_{1},E_{2},..,E_{N}}$\}. The performance is measured with the ability to take driving actions (e.g. steering) in a correct manner defined by the demonstrated behavior. 

To ensure and test generalization, the set of environments are split into training and testing environments; $E_{train}$  and $E_{test}$. The learning algorithm is allowed to access the ground truth actions set of training environments $E_{train}$ , while it is not allowed to access the ground truth actions of the test environments $E_{test}$ . The learning is said to be generalized if it achieves the desired performance on the test environments $E_{test}$. The architecture is based on two learners:

\begin{figure*}
\centerline{\includegraphics[width=120mm]{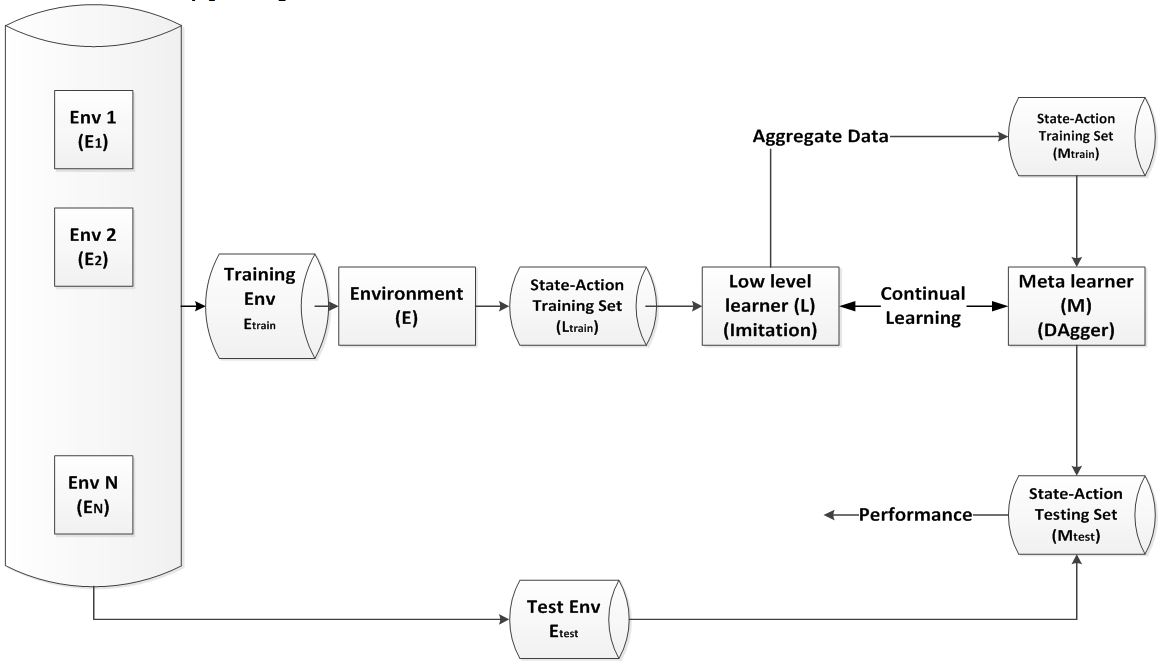}}
\caption{\label{Fig1} Meta Learning Framework for Dataset Aggregation}
\label{icml-historical}
\end{figure*} 

\subsection{Low Level Learner ($L$)}

The scope of the Low learner $L$ is to capture the specific features of each environment. The environment in operation $E$ is selected from a pool of training environments $E_{train}$. The low level learner is associated with a training data set $L_{train}$ which is formed due to interaction with an environment in operation $E$.  The training data set $L_{train}$  is formed of a list of state-action pairs \{${s,a}$\}$^t$, where t is the index of the interaction time stamp, with t=\{$0,1,...,T$\}, where $T$ is the end of interaction episode. An interaction episode is terminated with the $L$ reaching the goal, made a terminating mistake or after certain defined number of steps. The state $s$ is the measurement obtained by probing the environment $E$. The action $a$ is obtained from a demonstration, either from Human or from a reference algorithm.

Under the framework of Automated Driving, $L$ is trained using supervised learning imitating the demonstrated behavior.

\subsection{Meta Learner ($M$)}

The goal of the Meta Learner $M$ is to capture general features across all environments \{${E_{1},E_{2},..,E_{N}}$\}. With each interaction episode, $M$ communicates the initialization parameters to the Low learner $L$ to start interaction with the in operation environment $E$. After the Low learner finishes an episode, the learnt parameters of $L$ are communicated back to the $M$ to ensure continual learning.

The Meta Learner $M$ parameters are updated based on aggregated data over different interactions with the training environments $E_{train}$. The result of data set aggregation is called $M_{train}$. The members of $M_{train}$ have the same format of  state-action pairs  \{${s,a}$\}$^t$ as $L_{train}$. The training of $M$ is also following a supervised learning scheme using the training data in $M_{train}$.

The driving policy $\pi^M$ is obtained as the parameters of the Meta Learner $M$, such that an action at time $t$ is obtained as  . In the context of Neural Networks, $\pi^M$ is parameterized by the weights parameters of the network.
The performance of $\pi^M$ is tested against a test data set $M_{test}$, which is formed from the set of test environments $E_{test}$.

\section{MetaDAgger Algorithm} 
 
In this section the MetaDAgger algorithm is described as previous, in the light of the Meta learning framework described in $Figure$ \ref{Fig1}.

\begin{algorithm}[tb]
   \caption{MetaDAgger}
   \label{alg:example}
\begin{algorithmic}
   \STATE {\bfseries Loop on $E_{train}$:} // Data Collection Step
    \STATE  \qquad Collect $L_{train}$ += \{${s, a_{ref}}\}^t$ // $a_{ref}$ is obtained 
    \STATE  \qquad from a demonstration or a reference algorithm 
    \STATE  \qquad Aggregate $M_{train}$ += $L_{train}$
    \STATE Fit a Meta model $M$ based on $M_{train}$\\
    -------------------------------------------------------------------
   \STATE {\bfseries Loop until $N$-iter is reached} // Data Aggregation Step
   \STATE \qquad { \textbf{for} each $E$ in $E_{train}$} \textbf{do}
   \STATE \qquad \quad Initialize $L$ = $M$ 
   \STATE \qquad \quad Loop for $N_{steps}$
   \STATE \qquad \qquad $L$ interacts with $E$ to measure current state $s$
   \STATE \qquad \qquad Execute $a_{L}$ = $\pi^L$(s), where $\pi^{L}$ is obtained us-
   \STATE \qquad \qquad ing the parameters of $L$.
   \STATE \qquad \qquad \textbf{if}  {$a_{L} != a_{ref}$}: //Aggregate incorrect actions
   \STATE  \qquad \qquad \qquad Aggregate $L_{train}$ += \{${s, a_{ref}}$\}$^t$
    \STATE \qquad \qquad Fit a Low model $L$ based on $L_{train}$
    \STATE \qquad \quad Save back M=L
 \STATE Return $M$, $\pi^M$

\end{algorithmic}
\end{algorithm}

The algorithm is based on two main steps:

\subsection{Data Collection}
During data collection the reference demonstration is interacting with the environment to collect reference data. The reference demonstration could be a human or a reference algorithm. The collected data take the shape of state-action \{${s, a_{ref}}\}^t$ pairs at each interaction time step $t$, where $a_{ref}$  is the action provided by the demonstration.
This data is aggregated into $M_{train}$  to train the initial model parameters $M$, which could be a ConvNet fitting the supervised data.

\subsection{Data Aggregation}
Data aggregation is repeated for a number of iterations $N$-iter $>$ $|E_{train}|$, which means that the aggregation is repeated more than once over all the training environments.
Every iteration, $L$ interacts with $E$, starting from the parameters captures in $M$, which ensures continual learning. The parameters of $L$ define the interaction policy $\pi$ that will be executed. The interaction happens for N-steps, during which data aggregation takes place. The action to be executed is obtained from the policy parameters $\pi^{L}$ based on the measured state st. The aggregated data $L_{train}$ is then used to retrain $L$. At the end of each episode, the model is saved back to $M$ to ensure lifelong and continual learning for the next environment.

In order to improve sample efficiency, only the incorrect actions are saved. Hence, only actions that are different from the reference actions are saved. The difference is taken within certain tolerance (say 40\% error). This is important in two aspects: 1) the aggregated data set is focused on mistakes only, which need to be corrected when $L$ is re-trained, and 2) in real world, we have no access to the reference action $a_{ref}$, however, it is easier to ask a human supervisor to attend and correct only the algorithm mistakes, even before it causes damage, in which case the corrected state-action pair is aggregated and retraining happens to correct it.

\section{Experimental Setup} 

\subsection{Environment}
MetaDAgger is designed to be ready for deployment in real car. The algorithm steps ensure data sampling efficiency and zero damage to the experimental car. Moreover, aggregation of incorrect actions ensures easy human supervision requiring the least effort.

\begin{figure}
\vskip 0.2in
\begin{center}
\centerline{\includegraphics[width=50mm]{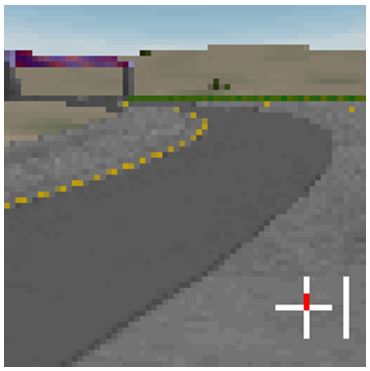}}
\caption{\label{Fig2}TORCS visual input state}
\label{icml-historical}
\end{center}
\vskip -0.2in
\end{figure}

As a first step towards deploying MetaDAgger in real car, we perform our experiments under TORCS game engine. We divide the 19 tracks in TORCS into 10 training tracks ($E_{train}$), and 9 test tracks ($E_{test}$). We use the Gym-TORCS environment \cite{GYM_TORCS}, with the visual image input \cite{Visual_TORCS}. Hence, the input state is the raw image pixels (64x64) as provided by the visual TORCS client as shown in $Figure$ \ref{Fig2}.
The actions provided by MetaDAgger are the continuous steering angle values. The supervision signal is obtained using a reference PID controller that can access the position of the car from the two side lanes, which is provided by the TORCS game engine. Although we could use this supervision signal to aggregate all encountered states in the data aggregation step, however, we keep aggregating only the incorrect ones. Since the actions are continuous, mapping the absolute action value to a steering angle is sensitive to small differences. Because of this, we relax the exact matching between the reference and algorithm actions to a certain tolerance, empirically set to 40\%.

\subsection{Network Architecture and Hyper Parameters}

\begin{figure*}
\centerline{\includegraphics[width=\textwidth]{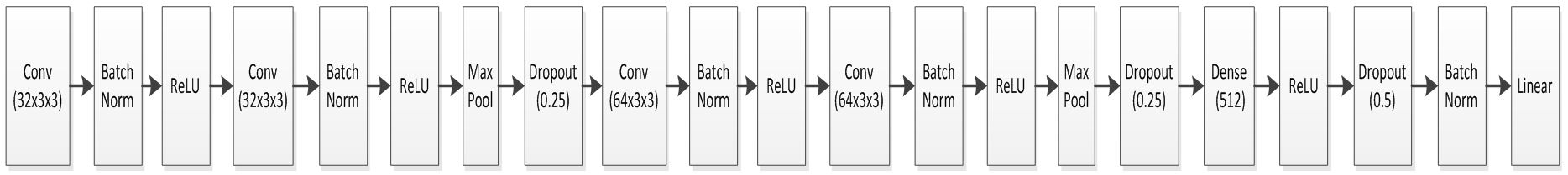}}
\caption{\label{Fig3}ConvNet architecture for Low and Meta learners}
\label{icml-historical}
\end{figure*}

For the Low learner $L$ and the Meta learner $M$ we fit the same ConvNet model shown in $Figure$ \ref{Fig3} , which simplifies copying models back and forth between $M$ and $L$ for continual learning. The kernel sizes are kept to a small size (3x3) due to the small input image size (64x64). Batch normalization is found to significantly improve the learning time \cite{ioffe2015batch}. Xaviar initialization \cite{glorot2010understanding} is used to initialize all weights. Dropout of 0.25 is used in the convolution layers, followed by Dropout of 0.5 in the fully connected ones.
ReLU activation is used in all layers, except for the output, where linear activation to produce the continuous output. The loss criterion is Mean Squared Error (MSE) minimization, since our task is a regression one.

\subsection{Results}
\begin{figure}
\begin{center}
\centerline{\includegraphics[width=\columnwidth]{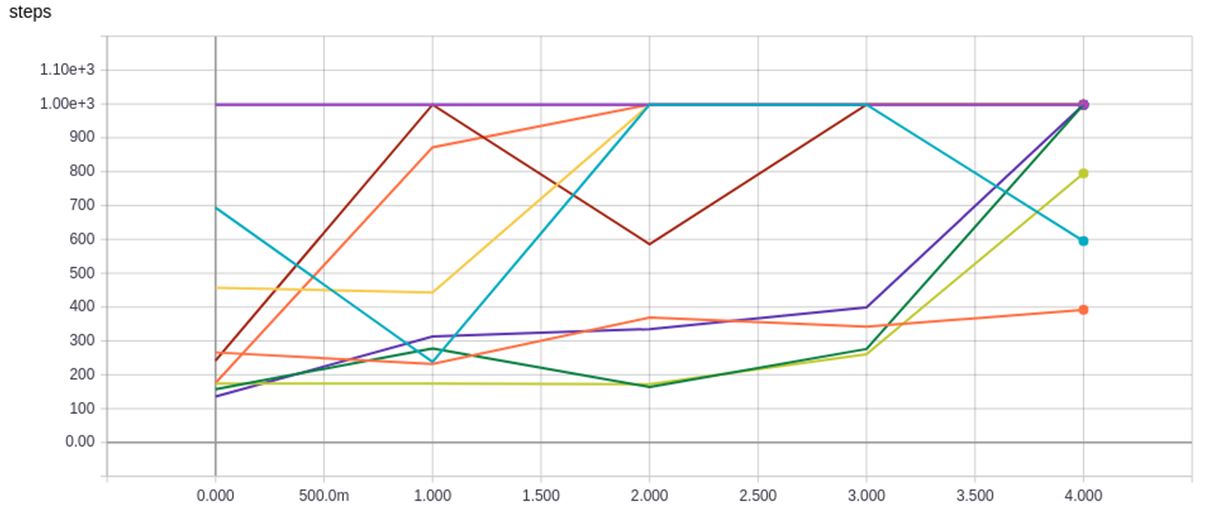}}
\caption{\label{Fig4}MetaDAgger performance on test tracks. Most of the tracks are completed just after 0 or 1 iterations of data aggregation.}
\label{icml-historical}
\end{center}
\end{figure}

We first evaluate the generalization performance of MetaDAgger. The evaluation is done over the 9 test tracks $M_{test}$. Results are shown in $Figure$ \ref{Fig4}. The horizontal axis represents the number of data aggregation iterations ($N$-iter), while the vertical axis is the number of steps the car moved without exiting the lane. We should note that 1000 steps mean one complete lap.

\begin{figure}
\begin{center}
\centerline{\includegraphics[width=\columnwidth]{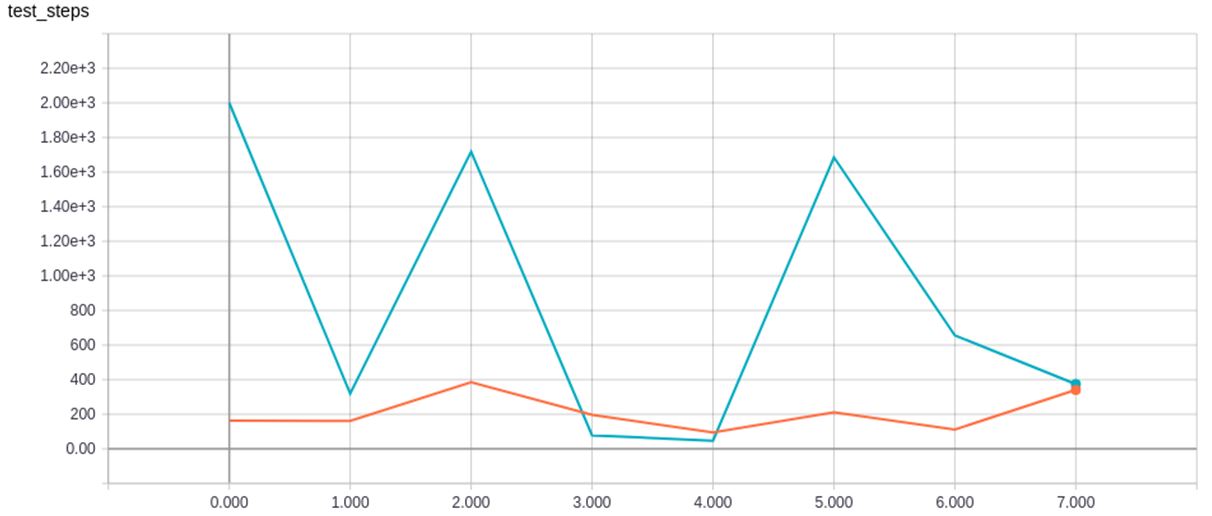}}
\caption{\label{Fig5}MetaDAgger vs. DAgger. Overall MetaDAgger is better than DAgger. For some tracks, it is still hard even with Meta learning to capture some hard turns.}
\label{icml-historical}
\end{center}
\end{figure}

We compare the performance of MetaDAgger against traditional DAgger \cite{ross2011reduction}. Results are shown in $Figure$ \ref{Fig5}, where the vertical axis represents the number of steps without collision (one complete lap equals 1000 steps), and the horizontal axis represents the test track number. For some of the test tracks, MetaDAgger is able to complete one or two laps. For other tracks, performance is better than DAgger, although no laps are completed.

\begin{figure}
\begin{center}
\centerline{\includegraphics[width=\columnwidth]{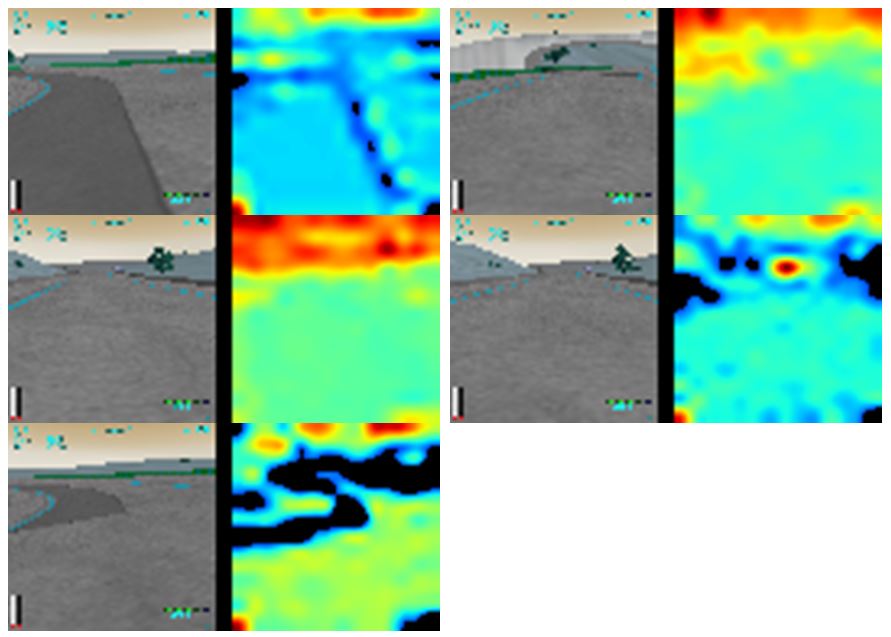}}
\caption{\label{Fig6}DAgger visualization. Most of the features represent the horizon or the mountain features.}
\label{icml-historical}
\end{center}
\end{figure}

We analyze the learnt features in case of DAgger and MetaDAgger. We use Grad CAM \cite{selvaraju2016grad}  visualization technique of the gradient of the last neuron (linear activation for action output), projected back to the input image. The result is a heat map showing which part of the image contributes more to the output. In case of DAgger, the learnt features are more memorizing and capturing the details of the track it is trained on, as shown in $Figure$ \ref{Fig6}. For example, we can see the most important part is the horizon or the place with the mountain; this is because the theme of this track is a desert one. It is then understood why it is hard for such model to generalize to other tracks with different themes.

\begin{figure}
\begin{center}
\centerline{\includegraphics[width=\columnwidth]{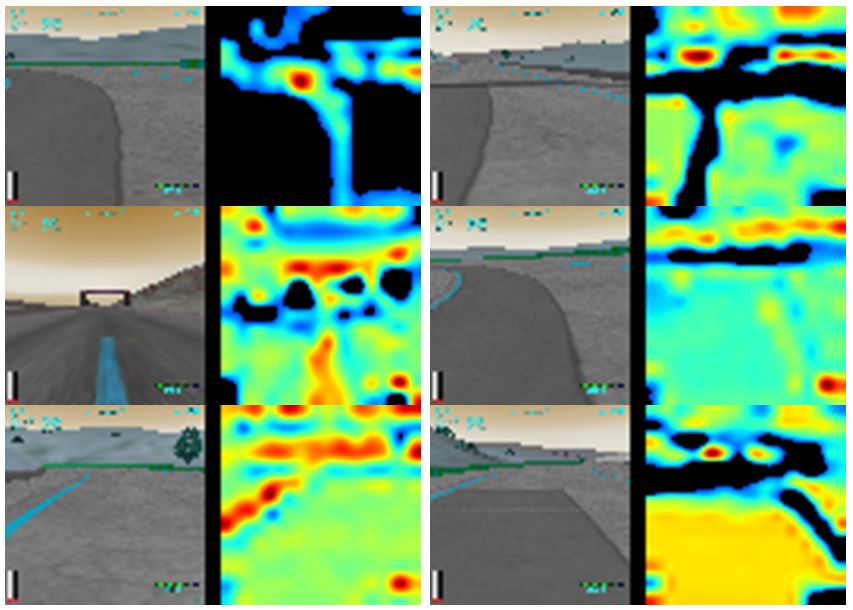}}
\caption{\label{Fig7}MetaDAgger filter visualization. Lane features are captured in many cases.}
\label{icml-historical}
\end{center}
\end{figure}

On the other hand, when we visualize the filters of MetaDAgger in $Figure$ \ref{Fig7}, we see that the learnt features are more representing the relevant features to the driving task, like the positions of the lanes. This proves the generality of such model.

\begin{figure}
\begin{center}
\centerline{\includegraphics[width=\columnwidth]{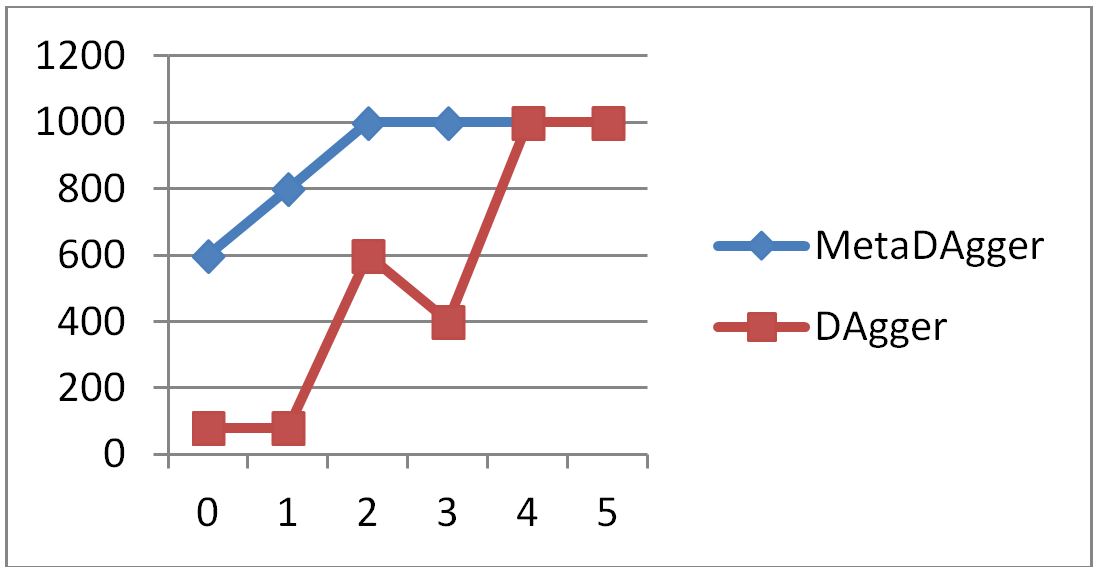}}
\caption{\label{Fig8}Sample efficiency of MetaDAgger vs. DAgger. After only data collection, MetaDAgger is able to complete 80\% of the lap.}
\label{icml-historical}
\end{center}
\end{figure}

We further evaluate the sample efficiency of MetaDAgger versus DAgger in $Figure$ \ref{Fig8}. Here we evaluate how many data aggregation iterations are needed in order for the algorithm to complete a lap. An iteration terminates after certain number of steps, a complete lap or a termination condition. In our experiments the number of steps are taken to be 1000 (complete lap), while the termination condition is met when the car is out of the track. In case of DAgger, it takes 4 iterations, while for MetaDAgger it takes only two. Moreover, just after the data collection and behavior cloning step, the algorithm is able to complete 80\% of the track.

\section{Conclusion}

In this paper we presented a framework for generalized imitation learning, based on the principles of Meta learning and data set aggregation. The proposed algorithm MetaDAgger is shown to be able to generalize on unseen test tracks, achieving much less training time and better sample efficiency.  The results on TORCS show significant improvement on both training and testing tracks, supported by visualizations of the generic learnt features. The proposed algorithm is designed to be ready for real car deployment, where the data aggregation step is only limited to correcting the mistakes the algorithm makes in real world.

\bibliography{example_paper}
\bibliographystyle{icml2017}

\end{document}